\documentclass[letterpaper, 10 pt, conference]{ieeeconf}  % Comment this line out if you need a4paper

\usepackage{graphicx}
\usepackage{multirow}
\usepackage{float}
\usepackage{amsmath}
\usepackage{amssymb,bm}
\usepackage{subcaption,xcolor}
\usepackage{color, soul}
\usepackage{url}
\usepackage{ctable}
\usepackage{pifont}

\IEEEoverridecommandlockouts                              % This command is only needed if
\title{\LARGE \bf
Real-Time Monocular Human Depth Estimation and Segmentation on Embedded Systems
}
\author{Shan An$^{1}$, Fangru Zhou$^{2}$, Mei Yang$^{2}$, Haogang Zhu$^{1,*}$, Changhong Fu$^{3}$,  and Konstantinos~A.~Tsintotas$^{4}$
\thanks{$^{1}$ Shan An and Haogang Zhu are with the School of Computer Science and Engineering, Beihang University, 100191 Beijing, China
        {\tt\small haogangzhu@buaa.edu.cn}}
\thanks{$^{2}$ Fangru Zhou and Mei Yang are with Tech \& Data Center, JD.COM Inc, 100108 Beijing, China
        {\tt\small zhoufangru@jd.com}
        }
\thanks{$^{3}$ Changhong Fu is with the School of Mechanical Engineering, Tongji University, 201804 Shanghai, China
        {\tt\small changhongfu@tongji.edu.cn}}
\thanks{$^{4}$ Konstantinos A. Tsintotas is with the Department of Production and Management Engineering, Democritus University of Thrace, 67132 Xanthi, Greece
        {\tt\small ktsintot@pme.duth.gr}}
\thanks{$^{*}$ Corresponding Author}
}

\begin{document}

\maketitle
\thispagestyle{empty}
\pagestyle{empty}

%%%%%%%%%%%%%%%%%%%%%%%%%%%%%%%%%%%%%%%%%%%%%%%%%%%%%%%%%%%%%%%%%%%%%%%%%%%%%%%%
\begin{abstract}

Estimating a scene's depth to achieve collision avoidance against moving pedestrians is a crucial and fundamental problem in the robotic field.
This paper proposes a novel, low complexity network architecture for fast and accurate human depth estimation and segmentation in indoor environments, aiming to applications for resource-constrained platforms (including battery-powered aerial, micro-aerial, and ground vehicles) with a monocular camera being the primary perception module. Following the encoder-decoder structure, the proposed framework consists of two branches, one for depth prediction and another for semantic segmentation. Moreover, network structure optimization is employed to improve its forward inference speed. Exhaustive experiments on three self-generated datasets prove our pipeline's capability to execute in real-time, achieving higher frame rates than contemporary state-of-the-art frameworks (114.6 frames per second on an NVIDIA Jetson Nano GPU with TensorRT) while maintaining comparable accuracy.

\end{abstract}

%%%%%%%%%%%%%%%%%%%%%%%%%%%%%%%%%%%%%%%%%%%%%%%%%%%%%%%%%%%%%%%%%%%%%%%%%%%%%%%%
\section{Introduction}

\begin{figure}[t]
	\begin{center}
   \includegraphics[width=1.0\linewidth]{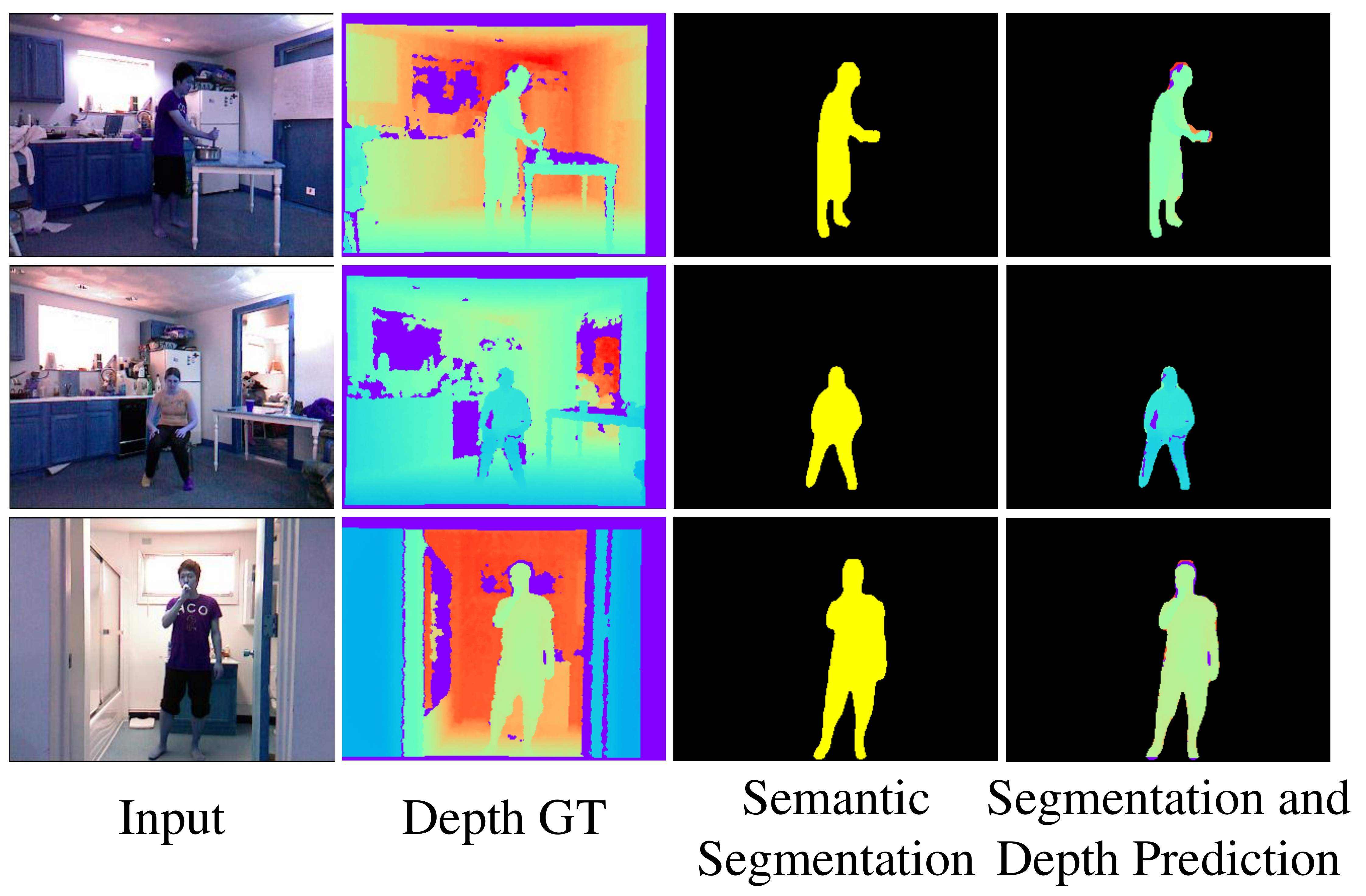} \par
	\end{center}
   \caption{Illustrative results of the proposed human depth estimation and segmentation framework.
		In the first column, the incoming RGB images are depicted.
		Since the encoder-decoder network structure consists of two branches, viz. depth prediction and semantic segmentation, the second column shows the metric depth data, while the third column presents the semantic segmentation results.
   The fourth column demonstrates the final segmentation and depth estimation of the foreground people instances.
   }
\label{fig:show}
\end{figure}

Depth estimation of a scene has been studied for a long time in the computer vision field for various applications, such as augmented reality \cite{diaz2017designing}, scene reconstruction \cite{Kusupati_2020_CVPR}, and detection \cite{mancini2016fast}.
In the robotic community, it is used for different tasks, which are mainly related to obstacle avoidance, localization, and mapping \cite{mancini2018j,chen2018collision}.
The ability of a robot to build a consistent map during its autonomous mission, widely known as Simultaneous Localization and Mapping (SLAM) \cite{cadena2016past}, is strengthened when scale information is provided as robust visual odometry is generated \cite{almalioglu2019ganvo}.
Thus, depth-sensing is essential in any contemporary SLAM system \cite{roussel2019monocular, mur2017orb}.
Commonly used sensors include LiDARs, binocular vision, etc., which are expensive and massive.
However, in most resource-constrained platforms (e.g. a micro aerial vehicle), cameras have become the primary perception device due to their low cost and power consumption.
As a result, in such cases, approaches that tackle the depth estimation task make use of a monocular camera.

Early studies were based on multi-scale features extracted from Convolutional Neural Networks (CNN) \cite{bhoi2019monocular}.
Firstly, they predicted coarse-scale depth information, and subsequently, they refined it through a fine-scaled network.
Current pipelines are mostly based on deep learning methods.
These are distinguished into three main categories, namely supervised, weakly-supervised, and unsupervised ones.
These frameworks adopt the encoder-decoder network structure \cite{durasov2019double,
%yang2019inferring,
hu2019revisiting, wang2020geom}, which originates from Natural Language Processing (NLP).
As the incoming camera stream arrives, the encoder extracts high-level, low-resolution features, while the decoder merges and upsamples them to produce the final high-resolution depth map.
Despite their high performances, these techniques are known for their excess demand in computational resources due to their high complexity functionality \cite{jing2020realtime, kansizoglou2020deep}.
Having identified this drawback, researchers developed frameworks with reduced computational complexity for real-time applications on embedded platforms \cite{nekrasov2019real, poggi2018towards, wofk2019fastdepth}.

In this paper, a straightforward network, which ensures real-time processing, for human depth estimation and segmentation is proposed.
The former is an essential information for obstacle avoidance, while the latter can permit the system to achieve more complex task simultaneously as it provides the crucial semantic data.
Our pipeline utilizes MobileNetV1 \cite{howard2017mobilenets} and Atrous Spatial Pyramid Pooling (ASPP) \cite{chen2018encoder} as the encoder, while the decoder is composed of depthwise separable convolutions and upsampling modules.
Furthermore, two branches, one for depth prediction and another for semantic segmentation, are proposed for the final estimation.
A network structure optimization is employed as well as the TensorRT optimizer \cite{tensorrt} to improve the forward inference speed.
An example containing results produced by our network is illustrated in Fig.~\ref{fig:show}.
%Moreover, to the best of our knowledge, there is no research regarding the simultaneous human depth estimation and segmentation using a monocular camera.
Various approaches \cite{durasov2019double, yang2019inferring, hu2019revisiting, wofk2019fastdepth, jafari2017analyzing} have been developed on the NYU Depth v2 dataset \cite{silberman2012indoor}, which is an indoor environment without humans, while the KITTI vision suite collection \cite{geiger2013vision} is selected for the outdoor cases \cite{roussel2019monocular, yang2019inferring, wang2020geom}.
% zhou2020windowed}.
Thus, no suitable data-sequence was available for our method's evaluation.
Therefore, to test the proposed framework, we generated three datasets based on the Cornell Activity \cite{CAD60} and the EPFL RGBD Pedestrian \cite{bagautdinov2015probability} image-sequences.
Utilizing the provided depth information and through the well-known segmentation method MaskRCNN \cite{he2017mask}, we automatically predicted the people masks, which subsequently were used as ground truth for the segmentation branch.
Finally, the proposed method is tested on these environments and compared against state-of-the-art approaches showing its improved performance.
An implementation of the presented work is available, under the title ``HDES-Net\footnote{https://github.com/AnshanTJU/HDES-Net}(Human Depth Estimation and Segmentation Network)".

The remainder of this work is structured as follows.
A literature review is presented in Section \ref{relatedwork}.
In Section \ref{method}, we describe our network design, whereas Section \ref{exper} evaluates and discusses the experimental results.
Finally, the last section is devoted to conclusions and future plans.

\section{Related Work}
\label{relatedwork}

\subsection{Monocular depth estimation}

Modern depth estimation methods use deep learning techniques trained over large-scale datasets.
Following the popular encoder-decoder structure, the authors in \cite{hu2019revisiting} propose a network with multi-scale feature fusion and refinement to produce accurate object boundaries.
A fast monocular depth estimation method is proposed by D. Wofk \textit{et al.}, which utilizes MobileNet as the encoder and depthwise decomposition in the decoder \cite{wofk2019fastdepth}.
This approach also utilizes the TVM compiler stack \cite{chen2018tvm} intending to address the runtime inefficiencies, while NetAdapt \cite{yang2018netadapt} is adopted for post-training network pruning.
Low complexity and low-latency are achieved, performing improved accuracy at 178 Frames Per Second (FPS) on an NVIDIA Jetson TX2 Graphics Processing Unit (GPU).
Except for the per-pixel depth, a pipeline can infer a distribution over possible depths through discrete binary classifications \cite{yang2019inferring}.
A double refinement network uses iterative pixel shuffle for upsampling \cite{shi2016real}.
In this method, the authors aim to replace the traditional bilinear interpolation and propose to guide the intermediate depth branch using auxiliary losses.
A geometric network is proposed in \cite{wang2020geom} to capture various structures of a scene, which is trained on uncalibrated videos.

\subsection{Semantic segmentation}

Semantic segmentation refers to the process that labels each pixel of an image with a corresponding class of what is represented.
Transferring and fine-tuning classification networks to Fully Convolutional Networks (FCN) \cite{long2015fully} show that improved performance can be achieved without further machinery.
Object interaction information is aggregated and fused to improve semantic segmentation performances \cite{bai2021information}.
Based on the common encoder-decoder architecture, U-Net \cite{ronneberger2015u} and SegNet \cite{badrinarayanan2017segnet} are widely used for semantic segmentation on medical \cite{ronneberger2015u} or satellite images \cite{ulku2020comparison}.
The well-known framework DeepLab series proposed Atrous convolution for dense feature extraction \cite{chen2017deeplab,chen2017rethinking}, ASPP \cite{chen2017deeplab} to encode objects, and a combination of CNN and fully-connected conditional random fields for accurate object boundary extraction \cite{chen2014semantic}.
DFANet \cite{li2019dfanet} utilizes a lightweight backbone and multi-scale feature propagation to reduce parameters.
This method exhibits sufficient performance and high inference speed.
Similarly, LEDNet \cite{wang2019lednet} proposes an asymmetric network for real-time semantic segmentation, while LiteSeg \cite{emara2019liteseg} explores ASPP to improve the segmentation results.
An Efficient Spatial Pyramid (ESP) module is proposed by ESPNet \cite{mehta2018espnet}, which uses a point-wise convolution and a pyramid of dilated convolutions to compose the final system.
In a later work, the same authors proposed ESPNetv2 \cite{mehta2019espnetv2}, where depthwise dilated separable convolutions are utilized to improve accuracy with fewer FLOPS.

%\subsection{Other studies}
%
%%While the aforementioned works focus either on depth estimation or semantic segmentation of a scene, there is only one study that leverages a similar task to ours.
%There is a study on 3D pedestrian localization with estimated uncertainty.
%L. Bertoni \textit{et al.} \cite{bertoni2019monoloco} use a human pose estimator to detect a set of keypoints and, afterwards, they fed them to a feed-forward neural network to predict the distance and the uncertainty associated with each pedestrian.
%Nevertheless, the multi-person pose estimation is computationally costly, while the method cannot segment humans from the scene, making it unsuitable for collision avoidance in robot applications.
%Our method is fundamentally different because it achieves real-time, low-complexity, monocular human depth estimation and segmentation at the same time.

\section{Proposed Method}
\label{method}

\begin{figure*}[ht]
\begin{center}
   \includegraphics[width=.95\linewidth]{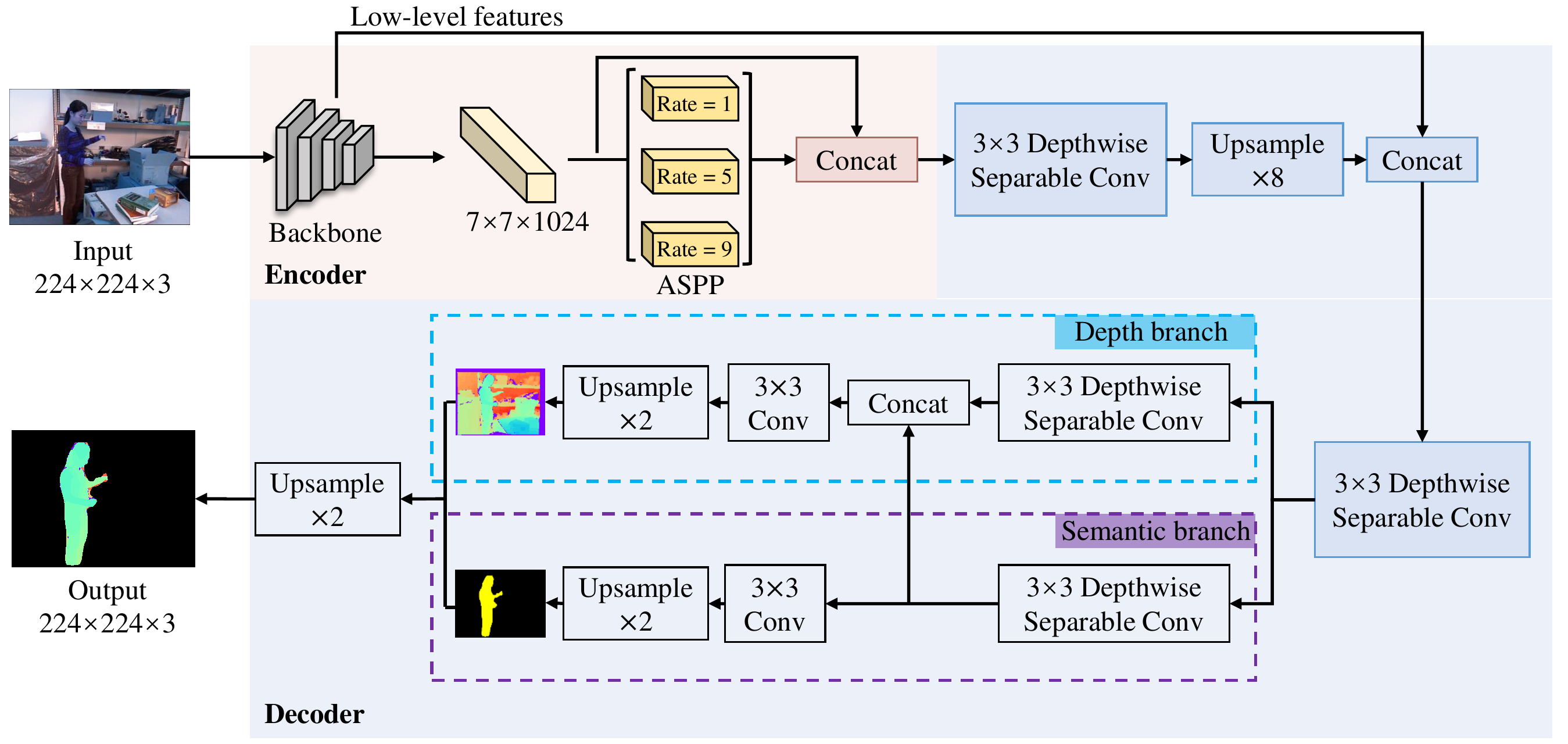} \par
\end{center}
   \caption{An overview of the proposed pipeline.
   MobileNetV1 \cite{howard2017mobilenets} is used as the network's backbone and combined with the Atrous Spatial Pyramid Pooling (ASPP) \cite{chen2018encoder} module to form the encoder.
   This way, high-level features with abstract information are extracted.
   In contrast, the decoder fuses the high-level features with the low-level details in order to predict the final estimation results through the depth and semantic branches.
   Aiming to intuitively understand and analyze the feature resolution of each stage of the network, the input size is $224\times224$.
	}
\label{fig:network}
\end{figure*}

In this section, we describe our network's design, which is a fully convolutional encoder-decoder network.
%A fast and accurate FCN architecture, both for the encoder and the decoder, is used.
The encoder extracts low-resolution, high-level abstract features from the provided visual sensory information, while through the decoder, a sufficiently high-resolution output is generated.
An outline of the proposed pipeline is depicted in Fig.~\ref{fig:network}.

\subsection{The encoder}

Commonly used networks, initially trained for image classification, such as VGG16 \cite{simonyan2014very} and ResNet-50 \cite{he2016deep}, have shown their improved capability to extract features with high accuracy.
As a result, they are usually selected as encoders.
However, despite their high performances, these approaches have two major disadvantages.
The first one is related to a large number of required computations, while the second one concerns the increased time of forward processing.
Therefore, as we aim for a real-time application which is able to work on embedded platforms, they are incompatible with our system.
MobileNetV1 \cite{howard2017mobilenets} is selected as the backbone of the encoder to achieve a balanced ratio between accuracy and processing time.
Furthermore, as the ASPP \cite{chen2018encoder} module has different sizes of receptive fields, making it capable of extracting features at different scales, we add it after the backbone in our network.
The depthwise separable convolutions are utilized to achieve lower execution times.
%we replaced the convolution layer with a depthwise and separable one to achieve lower execution times.
We use dilated convolutions \cite{yu2015multi} with a rate of $[1,3,6,9]$ in ASPP to increase the receptive field while maintaining the feature's resolution.

\subsection{The decoder}

As the encoder extracts features, the role of the decoder is to fuse and upsample them.
The ones that come from ASPP are merged with a stride of $4$.
This way, more detailed information is contained, which subsequently is used for depth estimation and semantic segmentation.
%Our decoder consists of two cascaded upsampling layers for expanding the feature's resolution by twice.
Our decoder consists of two upsampling layers, which upsample the feature maps eight times and twice, respectively.
Also, several depthwise separable layers perform $3\times3$ convolutions, reducing the number of output channels to $96$.
For the prediction step, two branches are proposed, i.e., one for depth estimation and another for semantic segmentation.
Both are composed of depthwise separable convolutions, standard convolutions, and upsampling layers.
%\hl{The depth of a scene is predicted for each pixel through batch normalization and activated by Relu.}
More specifically, the merged features are fed into the depthwise layer using a kernel size of $3$, stride length of $1$, and $96$ filters.
The output of this layer is upsampled twice to obtain the image's depth result, the size of which is half of the input frame.
Aiming to improve the performance, we add the features extracted from the depthwise separable convolutional layer of the semantic branch into the depth branch, as depicted in Fig. \ref{fig:network}.
Finally, a similar structure is adopted for the semantic segmentation pipeline where the output classes come from the standard convolution layer.
The loss of the semantic segmentation branch is smooth $L1$ loss, while the depth estimation branch utilizes cross-entropy loss.

\begin{table*}[h!]
	\caption{Properties of the used datasets. A ratio of $9:1$ is selected in order to divide the training and test set.}
	\label{table_dataset}
	\begin{center}
	\begin{tabular}{cl|c|c|c|c}
	\toprule
	\textbf{Dataset} & & \textbf{Description}  & \textbf{Image resolution} & \textbf{\# Training set} & \textbf{\# Test set} \\
	\midrule
	Cornell Activity \cite{CAD60}  & CAD-60 & Indoor, only one individual  & $240\times320$ & 74575 & 5737 \\
& CAD-120 &    & $480\times640$ & 60480 & 4653 \\
	\midrule
EPFL RGBD \cite{bagautdinov2015probability} &   & Lab and corridor, multiple pedestrians  & $424\times512$ & 4560 & 507 \\
	\bottomrule
	\end{tabular}
	\end{center}
\end{table*}

\subsection{Network structure optimization and acceleration}

A network structure optimization strategy is performed over ASPP branches to improve the forward inference speed.
Our main goal is to predict the depth and segment any human presenting in the input image within a distance of $10$ meters, which is the maximum depth of our datasets.
Therefore, the scale is relatively fixed, excluding the cases where the target is too large, exceeding the frame's covers.
Nevertheless, we retain the two parallel branches with dilation rates of $1$ and $9$, while we replace the remaining two with one presenting a dilation rate of $5$, as shown in Fig.~\ref{fig:network}.
%\hl{The network after pruning is trained and tested.}
Moreover, the global average pooling is removed and replaced with a dilated convolution of rate 5, which is more suitable for the human's scale in the image.
In addition to the above strategy, we also use TensorRT SDK \mbox{\cite{tensorrt}}, a deep learning inference optimizer, for further acceleration.

\section{Experimental Results}
\label{exper}

In this section, extensive experiments are conducted to demonstrate the effectiveness of the proposed architecture.
At first, we introduce our benchmark, the training settings, and the evaluation metrics.
Then, we provide ablation studies showing the effect of using the ASPP module, fusing the semantic information and network optimization.
Finally, through quantitative and qualitative experimentation, we measure the method's overall performance.

\subsection{Experimental settings}

\subsubsection{Benchmark introduction}

As there is no proper data-sequence related to simultaneous human depth estimation and segmentation in indoor scene, we generate three datasets based on the Cornell Activity \cite{CAD60} and EPFL RGBD \cite{bagautdinov2015probability}.
The former is composed of CAD-60 and CAD-120 image-sequences, which both contain RGB-D visual information of humans performing activities.
More specifically, CAD-60 has $60$ videos involving $4$ subjects with $12$ activities on $5$ different environments.
Regarding CAD-120, it consists of $120$ video of long daily activities involving $4$ subjects, $10$ high-level activities, $10$ sub-activity, and $12$ object affordance labels.
The camera measurements are recorded via the Microsoft Kinect sensor.
The EPFL RGBD Pedestrian dataset, which contains over 4000 RGB-D images, offers highly accurate depth maps thanks to a Kinect V2 module.
Table \ref{table_dataset} provides a brief description of each dataset used.
We divide the data into the training set and the test set with a ratio of $9:1$.
Subsequently, MaskRCNN \cite{he2017mask} trained on COCO dataset \cite{lin2014microsoft} was utilized to generate semantic segmentation masks.
We sample the annotations and verify them manually to ensure the correctness.
%Besides, along with the pedestrians' class, a set of another $57$ categories from MS COCO \cite{lin2014microsoft} is reserved.

\subsubsection{Training}

For CAD-60 and CAD-120 we used the SGD optimizer with $10^{-4}$ weight decay and $0.9$ momentum.
The initial learning rate was set to $10^{-2}$ and decayed to one-tenth of the previous one, while performed every $60$ epoch.
Regarding the EPFL RGBD Pedestrian data-sequence, we adopted the Adam optimizer with $5\times10^{-4}$ as weight decay.
The initial learning rate was set to $5\times10^{-4}$, while the decays to the half of the previous one as conducted every $100$ epochs.
The maximum number of iterations was $300$ epochs.
Network's implementation was made through the Pytorch framework \cite{paszke2019pytorch} utilizing a batch size of $64$, while an NVIDIA Tesla P40 GPU was used for the training procedure.
During network's training and testing, the images were not resized. Source code and some demo videos of the presented work can be found at \url{https://github.com/AnshanTJU/HDES-Net}.

\subsubsection{Evaluation metrics}

Three metrics are selected for evaluating the overall performance.
The RMSE (stands for Root Mean Squared Error in meters), $\delta_1$ (the percentage of predicted pixels where the relative error is within 25\%), and the People IoU (Intersection over Union).
The first two are chosen to evaluate the accuracy of human depth estimation, while the latter measures the semantic segmentation quality.

\subsection{Ablation studies}

\begin{table}[t!]
	\caption{The network's performance when Atrous Spatial Pyramid Pooling (ASPP) \cite{chen2018encoder} is applied.}
	\label{tab:aspp}
	\begin{center}
	\begin{tabular}{c|c|c|c}
	\toprule
	\textbf{Dataset} & \textbf{Metric} & \textbf{w/o ASPP} & \textbf{with ASPP} \\
	\midrule
	CAD-60 \cite{CAD60}  & RMSE $\downarrow$ 	 & 0.1529 	& \textbf{0.1526} \\
						 & $\delta_1$ $\uparrow$ & 98.71\% & \textbf{98.72\%} \\
						 & People IoU $\uparrow$ & 96.80\% & \textbf{96.82\%} \\
	\midrule	
	CAD-120 \cite{CAD60} & RMSE $\uparrow$ 	 	 & 0.3147  & \textbf{0.3140} \\
						 & $\delta_1$ $\uparrow$ & 99.97\% & \textbf{97.98\%} \\
						 & People IoU $\uparrow$ & 96.10\% & \textbf{96.17\%} \\
	\midrule
	EPFL RGBD \cite{bagautdinov2015probability}	 & RMSE $\downarrow$     & 0.1484  & \textbf{0.1461} \\
						 & $\delta_1$ $\uparrow$ & 98.47\% & \textbf{98.53\%} \\
						 & People IoU $\uparrow$ & \textbf{96.08\%} & 95.97\% \\	
	\bottomrule
	\end{tabular}
	\end{center}
\end{table}

\begin{table}[t!]
	\caption{\label{tab:fuse} {The network's performance when the features of the semantic branch are added into the depth branch.
}}
	\begin{center}
	\begin{tabular}{c|c|c|c}
	\toprule
	\textbf{Dataset} & \textbf{Metric} & \textbf{w/o fuse} & \textbf{fuse} \\
	\midrule
	CAD-60 \cite{CAD60}  & RMSE $\downarrow$ 	 & {0.1545} 	& \textbf{0.1526} \\
						 & $\delta_1$ $\uparrow$ & 98.69\% 			& \textbf{98.72\%} \\
						 & People IoU $\uparrow$ & {96.70\%}	& \textbf{96.82\%} \\
	\midrule	
	CAD-120 \cite{CAD60} & RMSE $\uparrow$ 	 	 & {0.3168}	& \textbf{0.3140} \\
						 & $\delta_1$ $\uparrow$ & 97.90\% 			& \textbf{97.98\%} \\
						 & People IoU $\uparrow$ & 95.89\% 			& \textbf{96.17\%} \\
	\midrule
	EPFL RGBD \cite{bagautdinov2015probability}	 & RMSE $\downarrow$     & 0.3185  			& \textbf{0.1461} \\
						 & $\delta_1$ $\uparrow$ & 97.91\% 			& \textbf{98.53\%} \\
						 & People IoU $\uparrow$ & {95.91\%} & \textbf{95.97\%} \\	
	\bottomrule
	\end{tabular}
	\end{center}
\end{table}

\begin{table}[t!]
	\caption{\label{tab:prun} {The network's performance through the utilization of the network structure optimization technique.
}}
	\begin{center}
	\renewcommand\tabcolsep{2pt}
	\begin{tabular}{c|c|c|c}
	\toprule
	\textbf{Dataset} & \textbf{Metric} & \textbf{w/o optimization} & \textbf{with optimization} \\
	\midrule
	CAD-60 \cite{CAD60}  & RMSE $\downarrow$ 	 & \textbf{0.1524} 	& 0.1526 \\
						 & $\delta_1$ $\uparrow$ & 98.72\% 			& \textbf{98.72\%} \\
						 & People IoU $\uparrow$ & \textbf{96.85\%}	& 96.82\% \\
	\midrule	
	CAD-120 \cite{CAD60} & RMSE $\uparrow$ 	 	 & \textbf{0.3133}	& 0.3140 \\
						 & $\delta_1$ $\uparrow$ & 97.98\% 			& \textbf{97.98\%} \\
						 & People IoU $\uparrow$ & 96.14\% 			& \textbf{96.17\%} \\
	\midrule
	EPFL RGBD \cite{bagautdinov2015probability}	 & RMSE $\downarrow$     & 0.1483  			& \textbf{0.1461} \\
						 & $\delta_1$ $\uparrow$ & 98.50\% 			& \textbf{98.53\%} \\
						 & People IoU $\uparrow$ & \textbf{96.13\%} & 95.97\% \\	
	\bottomrule
	\end{tabular}
	\end{center}
\end{table}

\begin{table}[t!]
	\caption{\label{tab:timeprun} {Measuring the inference runtime when the network structure optimization is employed.
	}}
	\resizebox{\columnwidth}{!}{
	\begin{tabular}{c|c|c}
	\toprule
	\textbf{Device} 					 & \textbf{w/o optimization} 	& \textbf{with optimization} \\
	\midrule
	Intel Xeon E5-2640 2.40GHz CPU 	     & 9.34 FPS  	 			& \textbf{13.80 FPS}  \\
	\midrule	
	NVIDIA Tesla P40 GPU   & 179.21	FPS	 			& \textbf{199.93 FPS} \\
	\midrule
	NVIDIA Jetson Nano GPU & 13.56 FPS	 			& \textbf{17.23 FPS} \\
	\bottomrule
	\end{tabular}
	}
\end{table}

\begin{table*}[h]
	\caption{\label{tab:backbone} {Comparing the inference runtime of our network when different backbones are used.}}
	\begin{center}
	\begin{tabular}{c|c|c|c|c|c}
	\toprule
	\textbf{Device} & \textbf{MobileNetV2 \cite{sandler2018mobilenetv2}} & \textbf{Resnet-18 \cite{he2016deep}} & \textbf{\textbf{Resnet-50 \cite{he2016deep}}} & \textbf{VGG16 \cite{simonyan2014very}} & \textbf{MobileNetV1 \cite{howard2017mobilenets}} \\
	\midrule
	Intel Xeon E5-2640 2.40GHz CPU		& 11.24 FPS  & 8.11 FPS & 6.49 FPS & 8.89 FPS   & \textbf{13.80 FPS} \\
	\midrule	
	NVIDIA Tesla P40 GPU & 120.22	FPS & 169.90 FPS & 112.91 FPS & 170.64 FPS & \textbf{199.93 FPS} \\
	\midrule
	NVIDIA Jetson Nano GPU & 14.30	FPS & 4.39 FPS & 3.46 FPS & 2.56 FPS & \textbf{17.23 FPS} \\
	\bottomrule
	\end{tabular}
	\end{center}
\end{table*}

\begin{table*}[t!]
	\caption{\label{tab:all} {Comparative results of the baseline methods against the proposed method. Green denotes the best, while blue is second best.}}	
	\begin{center}
	\begin{tabular}{c|c|c|c|c|c|c|c}
	\toprule
	\textbf{Dataset} & \textbf{Metric} & \textbf{LEDNet \cite{wang2019lednet}} & \textbf{LiteSeg \cite{emara2019liteseg}} & \textbf{ESPNet \cite{mehta2018espnet}} & \textbf{FastDepth \cite{wofk2019fastdepth}} & \textbf{DFANet \cite{li2019dfanet}} & \textbf{Our Proposed} \\
	\midrule
	CAD-60 \cite{CAD60}  & RMSE $\downarrow$ 	 & 0.2300 	 				& 0.1823   & \color{green}{0.1513}  & 0.1559   & 0.2833  & \color{blue}{0.1526} \\
					 	 & $\delta_1$ $\uparrow$ & 96.28\% 					& 98.25\%  & \color{green}{98.77\%} & 98.66\%  & 94.39\% & \color{blue}{98.72\%} \\
 					 	 & People IoU $\uparrow$ & \color{green}{97.31\%} 	& 94.35\%  & 95.06\% 				& 94.50\%  & 90.17\% & \color{blue}{96.82\%} \\
	\midrule	
	CAD-120 \cite{CAD60} & RMSE $\downarrow$ 	 & 0.4076 	 				& \color{green}{0.2982} & 0.3249  & 0.3323   	& 0.5259  				 & \color{blue}{0.3140} \\
					 	 & $\delta_1$ $\uparrow$ & 95.15\% 					& 98.12\%  				& \color{blue}{98.29\%} & \color{green}{98.31\%} & 88.78\% & 97.98\% \\
 					 	 & People IoU $\uparrow$ & \color{green}{96.64\%} 	& \color{blue}{96.51\%}  & 94.18\% & 94.48\%  	& 87.19\%  & 96.17\% \\
	\midrule
	EPFL RGBD \cite{bagautdinov2015probability}  & RMSE $\downarrow$ 	 	& 0.3882   & 0.2206   & \color{blue}{0.1804}  & 0.2663   & 0.8427  & \color{green}{0.1461} \\
					 	 & $\delta_1$ $\uparrow$ & 90.25\% 					& 96.33\%  & \color{blue}{97.70\%} & 94.07\%  & 50.92\% & \color{green}{98.53\%} \\
 					 	 & People IoU $\uparrow$ & \color{green}{96.50\%} 	& 93.96\%  & 90.75\% & 92.12\%  & 58.13\% & \color{blue}{95.97\%} \\
	\bottomrule
	\end{tabular}
	\end{center}
\end{table*}

\begin{table*}[h]
	\caption{\label{tab:alltime} {Inference runtime comparison between the baseline and the proposed network.}}
	\begin{center}
	\begin{tabular}{c|c|c|c|c|c|c}
	\toprule
	\textbf{Device} & \textbf{LEDNet \cite{wang2019lednet}} & \textbf{LiteSeg \cite{emara2019liteseg}} & \textbf{ESPNet \cite{mehta2018espnet}} & \textbf{FastDepth \cite{wofk2019fastdepth}} & \textbf{DFANet \cite{li2019dfanet}} & \textbf{Our Proposed} \\
	\midrule
	Intel Xeon E5-2640 2.40GHz CPU    & 13.66 FPS 	& 7.79 FPS & 17.65 FPS & \textbf{19.67 FPS}  & 6.79 FPS & {13.80 FPS} \\
	\midrule	
	NVIDIA Tesla P40 GPU & 70.73 FPS 	& 119.37 FPS & 133.09 FPS & 189.87 FPS  & 39.56 FPS & \textbf{199.93 FPS} \\
	\midrule
	NVIDIA Jetson Nano GPU & 4.39 FPS 	& 12.38 FPS & 11.42 FPS & 8.69 FPS  & 7.18 FPS & \textbf{17.23 FPS} \\
	\bottomrule
	\end{tabular}
	\end{center}
\end{table*}

\begin{figure*}[h!]
     \centering
     \begin{subfigure}[b]{0.35\linewidth}
         \centering
         \includegraphics[width=\textwidth]{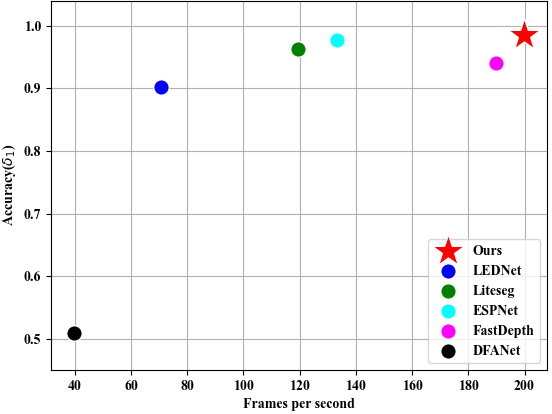}
         \caption{NVIDIA P40 GPU}
         \label{fig:speed1}
     \end{subfigure}
 ~
     \begin{subfigure}[b]{0.35\linewidth}
         \centering
         \includegraphics[width=\textwidth]{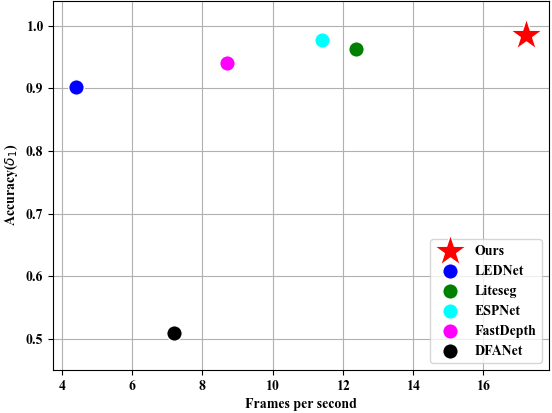}
         \caption{NVIDIA Jetson Nano GPU}
         \label{fig:speed2}
     \end{subfigure}
     \caption{The accuracy ($\delta_1$) and the runtime (measured in frames per second) when applied on NVIDIA P40 GPU (left) and NVIDIA Jetson Nano GPU (right) for various depth estimation frameworks.
    The EPFL \cite{bagautdinov2015probability} dataset is selected for evaluation.
	The input images are resized to $224\times224$.}
     \label{fig:speeds}
\end{figure*}

The ASPP module uses multiple dilated convolution branches with different rates to extract features at various scales.
This way, it can provide a better representation of humans in the image.
As shown in Table~\ref{tab:aspp}, where we compare our network's performance under the ASPP inclusion, an improvement is observed in almost every metric.

%In our network, the features extracted from the depthwise separable convolutional layer of the semantic branch are added into the depth branch.
%In Table~\ref{tab:fuse}, we show the effect of adding the semantic information into the depth branch.
%As can be seen, for the EPFL dataset with multiple pedestrians,  semantic segmentation information can guide the depth estimation branch to better distinguish the depth of different pedestrians.
%Therefore, whether the semantic branch information is fused or not has a great impact on the EPFL dataset.
%For CAD-60 and CAD-120 datasets, which contain a single person, semantic segmentation information is not as effective as in EPFL for optimizing depth.
%However, it can be seen that semantic information is still helpful for depth estimation.
Recall that we propose to incorporate the features from the semantic branch into the depth branch, the effect of this feature fusion operation is shown in Table~\ref{tab:fuse}. The results show that feature fusion indeed brings performance gains on all three benchmarks. One can draw that semantic segmentation information can lead to more precise depth estimation. Especially in the EPFL dataset with multiple pedestrians in each image, the incorporation of pedestrians' segmentation information helps the depth estimation branch to better distinguish the depth of each pedestrian, thus significantly boost the depth estimation performance.

In Table~\ref{tab:prun}, a performance comparison is presented, aiming to show the impact of the network structure optimization.
%As can been seen, there is no significant loss in precision, even if, in three metrics, a decreased score is achieved.
%Even if some of the three metrics are reduced, the overall performance is not decreased, and the algorithm with optimized structure performs better in many metrics.
Even if some of the three metrics are reduced, the overall performance is not decreased. The network with optimized structure shows a improved performance.
Also, we compare the network's forward processing speed for both cases, i.e., when the network structure optimization is employed and without it.
As shown in Table~\ref{tab:timeprun}, timings, measured in FPS, on different devices are significantly improved after optimization.
Finally, Table~\ref{tab:backbone} compares the proposed network when different backbones are applied.
As we can see, MobileNetV1 is the fastest one achieving nearly 200 FPS on a Tesla P40 GPU and 17.23 FPS on a Jetson Nano GPU.

\subsection{Comparison with the baseline techniques}

\begin{figure*}[h!]
	\begin{center}
   	\includegraphics[width=0.90\linewidth]{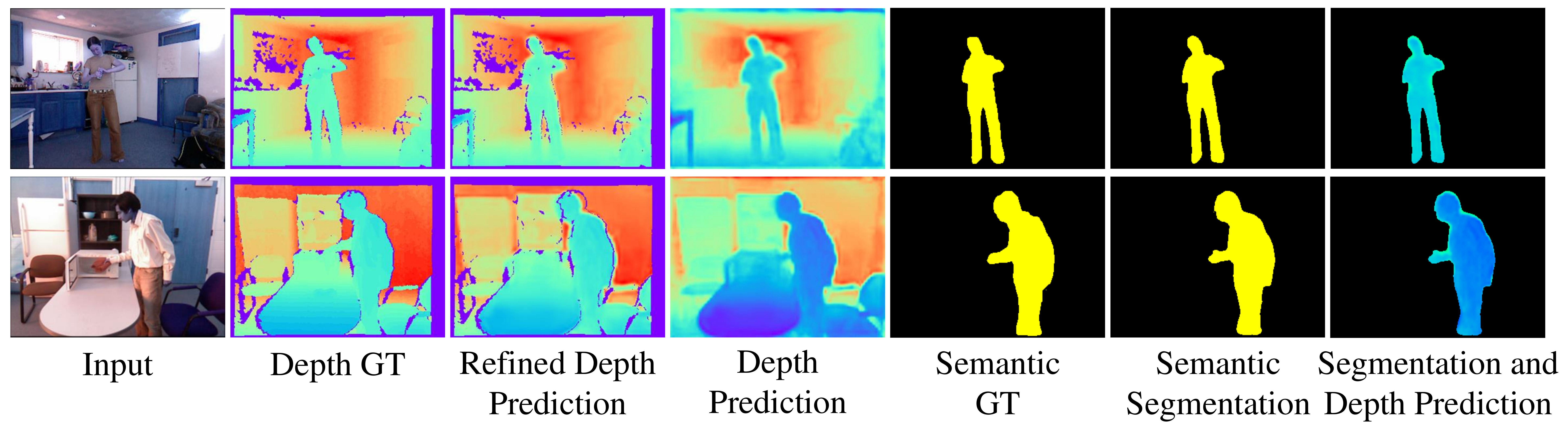}
	\end{center}
   	\caption{Qualitative results of CAD-60 dataset \cite{CAD60} (top) and CAD-120 dataset \cite{CAD60} (bottom).
   	}
	\label{fig:qua_cad}
\end{figure*}

\begin{figure*}[h!]
	\begin{center}
   	\includegraphics[width=0.90\linewidth]{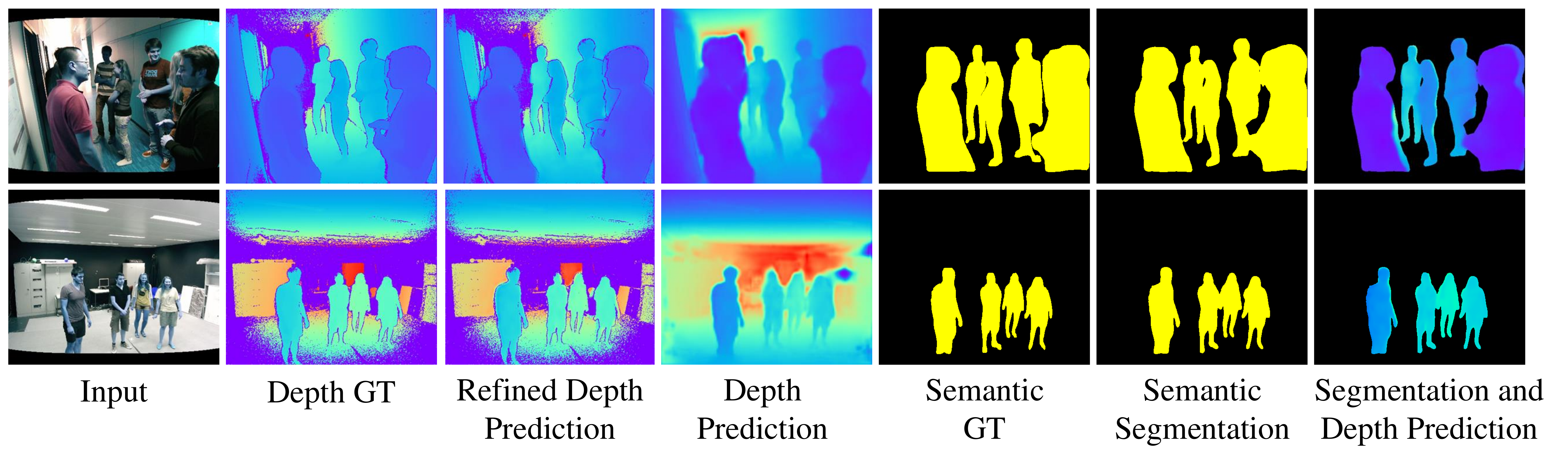}
	\end{center}
   	\caption{Qualitative results for the proposed pipeline on EPFL RGBD dataset \cite{bagautdinov2015probability}.}
	\label{fig:qua_epfl}
\end{figure*}

This section compares our method against other representative pipelines, which are: LEDNet \cite{wang2019lednet}, LiteSeg \cite{emara2019liteseg}, ESPNet \cite{mehta2018espnet}, FastDepth \cite{wofk2019fastdepth} and DFANet \cite{li2019dfanet}.
Each of the methods mentioned above add a depth estimation branch for the joint prediction.
In Table~\ref{tab:all}, we list the results obtained for each baseline method and the proposed network.
Green values indicate the highest scores, while the blue ones denote the second highest.
Since we aim to humans as the main object, People IoU is selected as a metric to demonstrate the performance regarding the semantic segmentation.
By examining Table~\ref{tab:all}, one can observe the significantly high scores achieved by our method in every evaluated dataset.
We succeed to excel among every other approach concerning the depth estimation on EPFL RGBD.
However, our framework performs unfavourably against other pipelines when compared on CAD-120.
The reason is that each dataset has a different maximum value of depth, which is 9.757, 12.4, and 8, for CAD-60, CAD-120, and EPFL RGBD, respectively.
Our algorithm is mainly developed to estimate depth in indoor scenes.
As a result, the accuracy of this value in a small range is relatively high.
Compared to EPFL RGBD, CAD-120 has a broader depth range, so our framework does not have a significant advantage over CAD-120.

Table~\ref{tab:alltime} compares the execution times needed for the proposed network and the baseline approaches when employed on different devices.
Notice the increased reduction offered by our network reaching a score of 199.93 FPS and 17.23 FPS on a Tesla P40 GPU and a Jetson Nano GPU, respectively.
As a final note, in Fig. \ref{fig:speed1} and Fig. \ref{fig:speed2}, our system is measured against other contemporary state-of-the-art solutions on the EPFL RGBD dataset through the accuracy $\delta_1$ over the processing speed (FPS).
It is noteworthy that our method outperforms each baseline approach.
ESPNet \cite{mehta2018espnet} and LiteSeg \cite{emara2019liteseg} achieve similar high scores regarding accuracy. Nevertheless, they are much slower.
We also test our network optimized with TensorRT on a Jetson Nano GPU.
The inference runtime comparison is shown in \mbox{Table~\ref{tab:tensorRT}}.
We can see that the model achieves 114.16 FPS, which far exceeds the real-time requirements.

Qualitative results of the proposed network are illustrated in Fig.~\ref{fig:qua_cad} and Fig.~\ref{fig:qua_epfl}, where illustrative results show that our method can accurately segment humans and estimate their depth.
There are some pixels with ignored depth values in images of the datasets.
We check the ground truth information and get these pixels and, then, we assign these pixels to zero to generate the refined depth prediction results.
In this way, it can be visually compared with ground truths more intuitively.

\begin{table}[]
	\caption{Measuring the inference runtime when TensorRT \cite{tensorrt} is applied.}
	\label{tab:tensorRT}
	\resizebox{\columnwidth}{!}
	{
	\begin{tabular}{c|c|c}
	\toprule
	\textbf{Device} & \textbf{w/o TensorRT} 		 & \textbf{with TensorRT} \\	
	\midrule
	NVIDIA Jetson Nano GPU & 17.23 FPS	 & \textbf{114.16 FPS} \\
	\bottomrule
	\end{tabular}
	}
\end{table}

\section{Conclusions and Future Work}
\label{conc}

As dense metric data allow a mobile robot to perform different tasks, such as obstacle avoidance and metric planning, to achieve a fully autonomous mission, in this paper, a real-time human depth estimation and segmentation network is proposed.
Our approach relies on the information provided by a monocular camera, while adopts computational low deep learning techniques to execute in real-time.
MobileNetV1, along with ASPP, is used to extract features at different scales, then fused and upsampled.
This way, we ensure high accuracy scores, while the processing speed is accelerated through network structure optimization and TensorRT optimizer reaching 114.16 FPS on a Jetson Nano GPU.
Our network is evaluated on three self-generated datasets demonstrating an improved performance compared to several state-of-the-art methods.
In our plans, we aim to use a monocular camera to realize learning-based collision avoidance in crowds.

% Given that our method is faster and has a smaller model, we can conclude that it is better than FastHand.

\section{Acknowledgement}
This work was supported by grants from the National Key Research and Development Program of China (Grant No. 2020YFC2006200).

%%%%%%%%%%%%%%%%%%%%%%%%%%%%%%%%%%%%%%%%%%%%%%%%%%%%%%%%%%%%%%%%%%%%%%%%%%%%%%%%

%%%%%%%%%%%%%%%%%%%%%%%%%%%%%%%%%%%%%%%%%%%%%%%%%%%%%%%%%%%%%%%%%%%%%%%%%%%%%%%%

%%%%%%%%%%%%%%%%%%%%%%%%%%%%%%%%%%%%%%%%%%%%%%%%%%%%%%%%%%%%%%%%%%%%%%%%%%%%%%%%
%\section*{APPENDIX}
%\section*{ACKNOWLEDGMENT}
%
%
%
%%%%%%%%%%%%%%%%%%%%%%%%%%%%%%%%%%%%%%%%%%%%%%%%%%%%%%%%%%%%%%%%%%%%%%%%%%%%%%%%%
%
%References are important to the reader; therefore, each citation must be complete and correct. If at all possible, references should be commonly available publications.

\bibliographystyle{IEEEtran}
\bibliography{ICRA-depth}

\begin{thebibliography}{10}
\providecommand{\url}[1]{#1}
\csname url@samestyle\endcsname
\providecommand{\newblock}{\relax}
\providecommand{\bibinfo}[2]{#2}
\providecommand{\BIBentrySTDinterwordspacing}{\spaceskip=0pt\relax}
\providecommand{\BIBentryALTinterwordstretchfactor}{4}
\providecommand{\BIBentryALTinterwordspacing}{\spaceskip=\fontdimen2\font plus
\BIBentryALTinterwordstretchfactor\fontdimen3\font minus
  \fontdimen4\font\relax}
\providecommand{\BIBforeignlanguage}[2]{{%
\expandafter\ifx\csname l@#1\endcsname\relax
\typeout{** WARNING: IEEEtran.bst: No hyphenation pattern has been}%
\typeout{** loaded for the language `#1'. Using the pattern for}%
\typeout{** the default language instead.}%
\else
\language=\csname l@#1\endcsname
\fi
#2}}
\providecommand{\BIBdecl}{\relax}
\BIBdecl

\bibitem{diaz2017designing}
C.~Diaz, M.~Walker, D.~A. Szafir, and D.~Szafir, ``Designing for depth
  perceptions in augmented reality,'' in \emph{Proc. IEEE Int. Symp. Mixed and
  Augmented Reality (ISMAR)}, 2017, pp. 111--122.

\bibitem{Kusupati_2020_CVPR}
U.~Kusupati, S.~Cheng, R.~Chen, and H.~Su, ``Normal assisted stereo depth
  estimation,'' in \emph{Proc. IEEE Conf. Computer Vision and Pattern
  Recognition (CVPR)}, June 2020.

\bibitem{mancini2016fast}
M.~Mancini, G.~Costante, P.~Valigi, and T.~A. Ciarfuglia, ``Fast robust
  monocular depth estimation for obstacle detection with fully convolutional
  networks,'' in \emph{Proc. IEEE/RSJ Int. Conf. Intelligent Robots and Systems
  (IROS)}, 2016, pp. 4296--4303.

\bibitem{mancini2018j}
------, ``{J-MOD 2: Joint monocular obstacle detection and depth estimation},''
  \emph{IEEE Robotics Automation Letters}, vol.~3, no.~3, pp. 1490--1497, 2018.

\bibitem{chen2018collision}
J.-H. Chen and K.-T. Song, ``Collision-free motion planning for human-robot
  collaborative safety under cartesian constraint,'' in \emph{Proc. IEEE Int.
  Conf. Robotics and Automation (ICRA)}, 2018, pp. 1--7.

\bibitem{cadena2016past}
C.~Cadena, L.~Carlone, H.~Carrillo, Y.~Latif, D.~Scaramuzza, J.~Neira, I.~Reid,
  and J.~J. Leonard, ``Past, present, and future of simultaneous localization
  and mapping: Toward the robust-perception age,'' \emph{IEEE Trans. robotics},
  vol.~32, no.~6, pp. 1309--1332, 2016.

\bibitem{almalioglu2019ganvo}
Y.~Almalioglu, M.~R.~U. Saputra, P.~P. de~Gusmao, A.~Markham, and N.~Trigoni,
  ``{GANVO}: Unsupervised deep monocular visual odometry and depth estimation
  with generative adversarial networks,'' in \emph{Proc. IEEE Int. Conf.
  Robotics and Automation (ICRA)}, 2019, pp. 5474--5480.

\bibitem{roussel2019monocular}
T.~Roussel, L.~Van~Eycken, and T.~Tuytelaars, ``Monocular depth estimation in
  new environments with absolute scale,'' in \emph{Proc. IEEE/RSJ Int. Conf.
  Intelligent Robots and Systems (IROS)}, 2019, pp. 1735--1741.

\bibitem{mur2017orb}
R.~Mur-Artal and J.~D. Tard{\'o}s, ``{ORB-SLAM2: An open-source slam system for
  monocular, stereo, and rgb-d cameras},'' \emph{IEEE Trans. Robotics},
  vol.~33, no.~5, pp. 1255--1262, 2017.

\bibitem{bhoi2019monocular}
A.~Bhoi, ``{Monocular depth estimation: A survey},'' \emph{arXiv preprint
  arXiv:1901.09402}, 2019.

\bibitem{durasov2019double}
N.~Durasov, M.~Romanov, V.~Bubnova, P.~Bogomolov, and A.~Konushin, ``Double
  refinement network for efficient monocular depth estimation.'' in \emph{Proc.
  IEEE/RSJ Int. Conf. Intelligent Robots and Systems (IROS)}, 2019, pp.
  5889--5894.

\bibitem{hu2019revisiting}
J.~Hu, M.~Ozay, Y.~Zhang, and T.~Okatani, ``{Revisiting single image depth
  estimation: Toward higher resolution maps with accurate object boundaries},''
  in \emph{Proc. IEEE Wint. Conf. Appli. Comput. Vision (WACV)}, 2019, pp.
  1043--1051.

\bibitem{wang2020geom}
K.~Wang, Y.~Chen, H.~Guo, L.~Wen, and S.~Shen, ``Geometric pretraining for
  monocular depth estimation,'' in \emph{Proc. IEEE Int. Conf. Robotics and
  Automation (ICRA)}, 2020, pp. 4782--4788.

\bibitem{jing2020realtime}
J.~Liang, U.~Patel, A.~J. Sathyamoorthy, and D.~Manocha, ``Realtime collision
  avoidance for mobile robots in dense crowds using implicit multi-sensor
  fusion and deep reinforcement learning,'' in \emph{Proc. Int. Conf.
  Autonomous Agents and Multi-Agents Systems (AAMAS)}, 2020.

\bibitem{kansizoglou2020deep}
I.~Kansizoglou, L.~Bampis, and A.~Gasteratos, ``Deep feature space: A
  geometrical perspective,'' \emph{arXiv preprint arXiv:2007.00062}, 2020.

\bibitem{nekrasov2019real}
V.~Nekrasov, T.~Dharmasiri, A.~Spek, T.~Drummond, C.~Shen, and I.~Reid,
  ``Real-time joint semantic segmentation and depth estimation using asymmetric
  annotations,'' in \emph{Proc. IEEE Int. Conf. Robotics and Automation
  (ICRA)}, 2019, pp. 7101--7107.

\bibitem{poggi2018towards}
M.~Poggi, F.~Aleotti, F.~Tosi, and S.~Mattoccia, ``Towards real-time
  unsupervised monocular depth estimation on cpu,'' in \emph{Proc. IEEE/RSJ
  Int. Conf. Intelligent Robots and Systems (IROS)}, 2018, pp. 5848--5854.

\bibitem{wofk2019fastdepth}
D.~Wofk, F.~Ma, T.-J. Yang, S.~Karaman, and V.~Sze, ``{FastDepth}: Fast
  monocular depth estimation on embedded systems,'' in \emph{Proc. IEEE Int.
  Conf. Robotics and Automation (ICRA)}, 2019, pp. 6101--6108.

\bibitem{howard2017mobilenets}
A.~G. Howard, M.~Zhu, B.~Chen, D.~Kalenichenko, W.~Wang, T.~Weyand,
  M.~Andreetto, and H.~Adam, ``{MobileNets}: Efficient convolutional neural
  networks for mobile vision applications,'' \emph{arXiv preprint
  arXiv:1704.04861}, 2017.

\bibitem{chen2018encoder}
L.-C. Chen, Y.~Zhu, G.~Papandreou, F.~Schroff, and H.~Adam, ``Encoder-decoder
  with atrous separable convolution for semantic image segmentation,'' in
  \emph{Proc. Eur. Conf. Computer Vision (ECCV)}, 2018, pp. 801--818.

\bibitem{tensorrt}
\BIBentryALTinterwordspacing
(2020) {TensorRT Open Source Software}. [Online]. Available:
  \url{https://github.com/NVIDIA/TensorRT}
\BIBentrySTDinterwordspacing

\bibitem{yang2019inferring}
G.~Yang, P.~Hu, and D.~Ramanan, ``Inferring distributions over depth from a
  single image,'' \emph{arXiv preprint arXiv:1912.06268}, 2019.

\bibitem{jafari2017analyzing}
O.~H. Jafari, O.~Groth, A.~Kirillov, M.~Y. Yang, and C.~Rother, ``{Analyzing
  modular CNN architectures for joint depth prediction and semantic
  segmentation},'' in \emph{Proc. IEEE Int. Conf. Robotics and Automation
  (ICRA)}, 2017, pp. 4620--4627.

\bibitem{silberman2012indoor}
N.~Silberman, D.~Hoiem, P.~Kohli, and R.~Fergus, ``Indoor segmentation and
  support inference from rgbd images,'' in \emph{Proc. Euro. Conf. Compututer
  Vision (ECCV)}, 2012, pp. 746--760.

\bibitem{geiger2013vision}
A.~Geiger, P.~Lenz, C.~Stiller, and R.~Urtasun, ``{Vision meets robotics: The
  kitti dataset},'' \emph{Int. J. Robotics Research}, vol.~32, no.~11, pp.
  1231--1237, 2013.

\bibitem{CAD60}
\BIBentryALTinterwordspacing
(2009) Cornell activity datasets: {CAD-60} \& {CAD-120}. [Online]. Available:
  \url{http://pr.cs.cornell.edu/humanactivities/data.php}
\BIBentrySTDinterwordspacing

\bibitem{bagautdinov2015probability}
T.~Bagautdinov, F.~Fleuret, and P.~Fua, ``Probability occupancy maps for
  occluded depth images,'' in \emph{Proc. IEEE Conf. Computer Vision and
  Pattern Recognition (CVPR)}, 2015, pp. 2829--2837.

\bibitem{he2017mask}
K.~He, G.~Gkioxari, P.~Doll{\'a}r, and R.~Girshick, ``Mask {R-CNN},'' in
  \emph{Proc. IEEE Int. Conf. Computer Vision (ICCV)}, 2017, pp. 2961--2969.

\bibitem{chen2018tvm}
T.~Chen, T.~Moreau, Z.~Jiang, L.~Zheng, E.~Yan, H.~Shen, M.~Cowan, L.~Wang,
  Y.~Hu, L.~Ceze \emph{et~al.}, ``{TVM}: An automated end-to-end optimizing
  compiler for deep learning,'' in \emph{Proc. 13th Symp. Operating Systems
  Design and Implementation (OSDI)}, 2018, pp. 578--594.

\bibitem{yang2018netadapt}
T.-J. Yang, A.~Howard, B.~Chen, X.~Zhang, A.~Go, M.~Sandler, V.~Sze, and
  H.~Adam, ``{NetAdapt}: Platform-aware neural network adaptation for mobile
  applications,'' in \emph{Proc. Eur. Conf. Computer Vision (ECCV)}, 2018, pp.
  285--300.

\bibitem{shi2016real}
W.~Shi, J.~Caballero, F.~Husz{\'a}r, J.~Totz, A.~P. Aitken, R.~Bishop,
  D.~Rueckert, and Z.~Wang, ``Real-time single image and video super-resolution
  using an efficient sub-pixel convolutional neural network,'' in \emph{Proc.
  IEEE Conf. Computer Vision and Pattern Recognition (CVPR)}, 2016, pp.
  1874--1883.

\bibitem{long2015fully}
J.~Long, E.~Shelhamer, and T.~Darrell, ``Fully convolutional networks for
  semantic segmentation,'' in \emph{Proc. IEEE Conf. Comp. Vision and Pattern
  Recognition (CVPR)}, 2015, pp. 3431--3440.

\bibitem{bai2021information}
S.~Bai and C.~Wang, ``Information aggregation and fusion in deep neural
  networks for object interaction exploration for semantic segmentation,''
  \emph{Knowledge-Based Systems}, vol. 218, p. 106843, 2021.

\bibitem{ronneberger2015u}
O.~Ronneberger, P.~Fischer, and T.~Brox, ``{U-Net}: Convolutional networks for
  biomedical image segmentation,'' in \emph{Proc. Int. Conf. Medical Image
  Computing and Computer-Assisted Intervention (MICCAI)}, 2015, pp. 234--241.

\bibitem{badrinarayanan2017segnet}
V.~Badrinarayanan, A.~Kendall, and R.~Cipolla, ``{SegNet}: A deep convolutional
  encoder-decoder architecture for image segmentation,'' \emph{IEEE Trans.
  Pattern Analysis and Machine Intelligence}, vol.~39, no.~12, pp. 2481--2495,
  2017.

\bibitem{ulku2020comparison}
I.~Ulku, P.~Barmpoutis, T.~Stathaki, and E.~Akagunduz, ``Comparison of single
  channel indices for u-net based segmentation of vegetation in satellite
  images,'' in \emph{Proc. 20th Int. Conf. Machine Vision (ICMV)}, 2020.

\bibitem{chen2017deeplab}
L.-C. Chen, G.~Papandreou, I.~Kokkinos, K.~Murphy, and A.~L. Yuille,
  ``{DeepLab}: Semantic image segmentation with deep convolutional nets, atrous
  convolution, and fully connected crfs,'' \emph{IEEE Trans. Pattern Analysis
  and Machine Intelligence}, vol.~40, no.~4, pp. 834--848, 2017.

\bibitem{chen2017rethinking}
L.-C. Chen, G.~Papandreou, F.~Schroff, and H.~Adam, ``Rethinking atrous
  convolution for semantic image segmentation,'' \emph{arXiv preprint
  arXiv:1706.05587}, 2017.

\bibitem{chen2014semantic}
L.-C. Chen, G.~Papandreou, I.~Kokkinos, K.~Murphy, and A.~L. Yuille, ``Semantic
  image segmentation with deep convolutional nets and fully connected crfs,''
  \emph{arXiv preprint arXiv:1412.7062}, 2014.

\bibitem{li2019dfanet}
H.~Li, P.~Xiong, H.~Fan, and J.~Sun, ``{DFANet}: Deep feature aggregation for
  real-time semantic segmentation,'' in \emph{Proc. IEEE Conf. Computer Vision
  and Pattern Recognition (CVPR)}, 2019, pp. 9522--9531.

\bibitem{wang2019lednet}
Y.~Wang, Q.~Zhou, J.~Liu, J.~Xiong, G.~Gao, X.~Wu, and L.~J. Latecki,
  ``{LEDNet}: A lightweight encoder-decoder network for real-time semantic
  segmentation,'' in \emph{Proc. IEEE Int. Conf. Image Processing (ICIP)},
  2019, pp. 1860--1864.

\bibitem{emara2019liteseg}
T.~Emara, H.~E. Abd El~Munim, and H.~M. Abbas, ``{LiteSeg}: A novel lightweight
  convnet for semantic segmentation,'' in \emph{Proc. Digital Image Computing:
  Techniques and Applications}, 2019, pp. 1--7.

\bibitem{mehta2018espnet}
S.~Mehta, M.~Rastegari, A.~Caspi, L.~Shapiro, and H.~Hajishirzi, ``{ESPNet}:
  Efficient spatial pyramid of dilated convolutions for semantic
  segmentation,'' in \emph{Proc. Eur. Conf. Computer Vision (ECCV)}, 2018, pp.
  552--568.

\bibitem{mehta2019espnetv2}
S.~Mehta, M.~Rastegari, L.~Shapiro, and H.~Hajishirzi, ``{ESPNetv2}: A
  light-weight, power efficient, and general purpose convolutional neural
  network,'' in \emph{Proc. IEEE Conf. Computer Vision and Pattern Recognition
  (CVPR)}, 2019, pp. 9190--9200.

\bibitem{simonyan2014very}
K.~Simonyan and A.~Zisserman, ``Very deep convolutional networks for
  large-scale image recognition,'' \emph{arXiv preprint arXiv:1409.1556}, 2014.

\bibitem{he2016deep}
K.~He, X.~Zhang, S.~Ren, and J.~Sun, ``Deep residual learning for image
  recognition,'' in \emph{Proc. IEEE Conf. Computer Vision and Pattern
  Recognition (CVPR)}, 2016, pp. 770--778.

\bibitem{yu2015multi}
F.~Yu and V.~Koltun, ``Multi-scale context aggregation by dilated
  convolutions,'' \emph{arXiv preprint arXiv:1511.07122}, 2015.

\bibitem{lin2014microsoft}
T.-Y. Lin, M.~Maire, S.~Belongie, J.~Hays, P.~Perona, D.~Ramanan,
  P.~Doll{\'a}r, and C.~L. Zitnick, ``{Microsoft COCO}: Common objects in
  context,'' in \emph{Proc. Eur. Conf. Computer Vision (ECCV)}, 2014, pp.
  740--755.

\bibitem{paszke2019pytorch}
A.~Paszke, S.~Gross, F.~Massa, A.~Lerer, J.~Bradbury, G.~Chanan, T.~Killeen,
  Z.~Lin, N.~Gimelshein, L.~Antiga \emph{et~al.}, ``Pytorch: An imperative
  style, high-performance deep learning library,'' in \emph{Advances in neural
  information processing systems}, 2019, pp. 8026--8037.

\bibitem{sandler2018mobilenetv2}
M.~Sandler, A.~Howard, M.~Zhu, A.~Zhmoginov, and L.-C. Chen, ``{MobileNetV2}:
  Inverted residuals and linear bottlenecks,'' in \emph{Proc. IEEE Conf.
  Computer Vision and Pattern Recognition (CVPR)}, 2018, pp. 4510--4520.

\end{thebibliography}

\end{document}